\documentclass[letterpaper, 10 pt, conference]{ieeeconf}
\IEEEoverridecommandlockouts
\usepackage{gensymb}
\usepackage{graphicx}
\usepackage{amsmath,amssymb}
\usepackage{cite}
\usepackage{xcolor}
\usepackage{multirow}
\usepackage{makecell}
\usepackage{tabularx}
\usepackage{url}
\usepackage{fancyhdr} 

\fancypagestyle{firstpage}{
    \fancyhf{} 
    \fancyfoot[C]{\footnotesize This work has been submitted to the IEEE for possible publication. Copyright may be transferred without notice, after which this version may no longer be accessible.}
}

\fancypagestyle{plain}{
    \fancyhf{} 
    \fancyfoot[C]{\footnotesize This work has been submitted to the IEEE for possible publication. Copyright may be transferred without notice, after which this version may no longer be accessible.}
}

\begin{document}

\title{\LARGE \bf
Learning Speed-Adaptive Walking Agent Using Imitation Learning with Physics-Informed Simulation}

\author{Yi-Hung Chiu*$^{1}$, Ung Hee Lee*$^{2}$, Changseob Song$^{1}$, Manaen Hu$^{1}$, and Inseung Kang$^{1}$ \textit{Member, IEEE}
\thanks{* Y. Chiu and U. Lee contributed equally to the paper. Corresponding author: {\tt\footnotesize unghee@umich.edu}.}
\thanks{$^{1}$Y. Chiu, M. Hu, C. Song, and I. Kang are with the Department of Mechanical Engineering, Carnegie Mellon University, Pittsburgh PA 15213, USA.
}
\thanks{$^{2}$U. Lee is with the Department of Mechanical Engineering and the Department of
Robotics, University of Michigan, Ann Arbor, MI 48109 USA.
}
}

\maketitle
\thispagestyle{plain}
\begin{abstract}
Virtual models of human gait, or digital twins, offer a promising solution for studying mobility without the need for labor-intensive data collection. However, challenges such as the sim-to-real gap and limited adaptability to diverse walking conditions persist. To address these, we developed and validated a framework to create a skeletal humanoid agent capable of adapting to varying walking speeds while maintaining biomechanically realistic motions. The framework combines a synthetic data generator, which produces biomechanically plausible gait kinematics from open-source biomechanics data, and a training system that uses adversarial imitation learning to train the agent’s walking policy. We conducted comprehensive analyses comparing the agent's kinematics, synthetic data, and the original biomechanics dataset. The agent achieved a root mean square error of 5.24$\pm$0.09 degrees at varying speeds compared to ground-truth kinematics data, demonstrating its adaptability. This work represents a significant step toward developing a digital twin of human locomotion, with potential applications in biomechanics research, exoskeleton design, and rehabilitation.
\end{abstract}


\begin{figure*}[!t]
    \centering
    \includegraphics[width=0.85\linewidth]{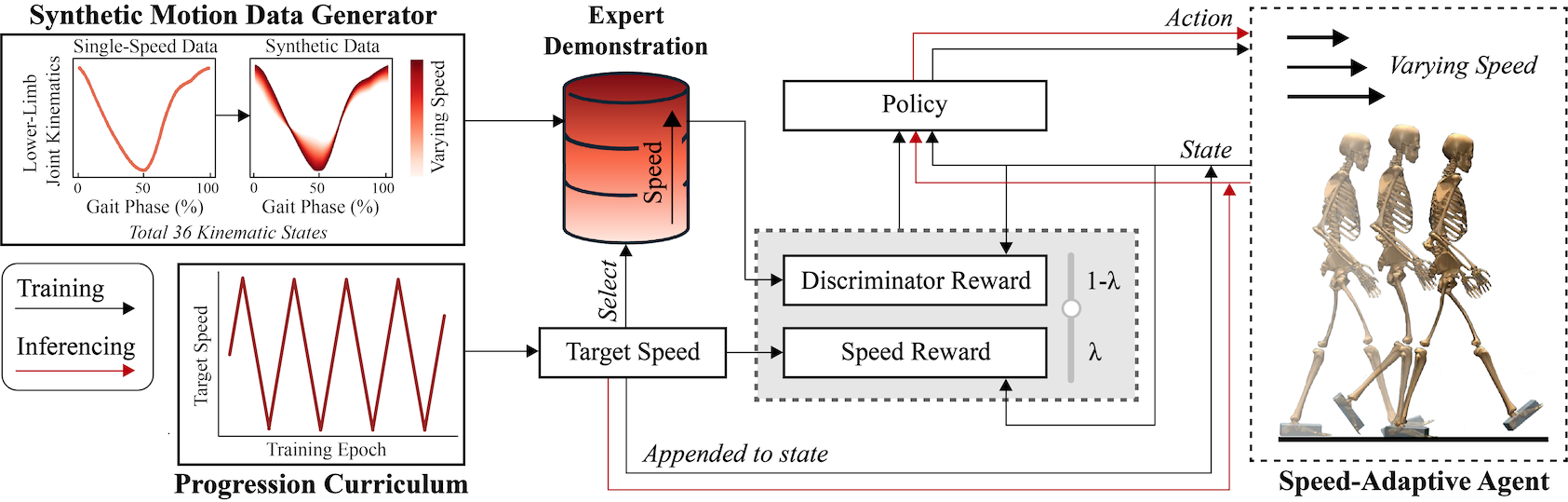}
    \caption{Overview of our framework for training a speed-adaptive walking agent. The framework includes two main components: generating expert demonstrations and policy optimization. Expert demonstrations from motion data are used to train a discriminator that distinguishes expert- from policy-generated actions at different speeds. The reward function combines the discriminator's output with a speed reward (difference between the agent's target and actual center of mass speed). These rewards optimize the walking policy using trust region policy optimization. The target speed is part of the agent's observation space, with a progressive curriculum exposing the agent to a range of speeds during training. Evaluation tests the agent's ability to achieve the desired speed under varying conditions.}
    \label{fig:overview}
\end{figure*}

\section{Introduction}
Human mobility is a fundamental aspect of daily life, as it allows essential activities, promotes health through physical exercise, and contributes to psychological well-being \cite{satariano2012mobility, webber2010mobility}. However, millions of people in the United States alone face mobility challenges due to conditions such as loss of limbs, paraplegia, and obesity, which decrease quality of life by affecting daily tasks, limiting exercise, and often leading to mental health issues such as depression and anxiety \cite{cdc2024disability, iezzoni2001mobility}. Restoring gait functionality is crucial to improving the quality of life of people with mobility impairments, necessitating a deep understanding of gait mechanics. Although traditional methods such as motion capture systems and force plates have been used for gait analysis \cite{baker2006gait, webber2010mobility}, these approaches are costly, time-consuming, and labor intensive, highlighting the need for more efficient and accessible alternatives to support rehabilitation and improve mobility outcomes. \par

Digital twins, a virtual representation of physical systems, have emerged as a powerful tool in biomechanics and rehabilitation \cite{delp2007opensim, barricelli2019survey}. These computational models simulate human gait biomechanics in controlled environments, eliminating the need to collect human motion data while providing valuable information on joint coordination, muscle activities, and gait balance \cite{delp1990interactive, ong2019predicting}. In addition, these models are particularly beneficial when developing wearable assistive devices or novel rehabilitation techniques \cite{sreenivasa2017optimal, kumar2020learning, tran2022lightweight, kang2019effect}. Digital twins can model how these devices interact with the human body, offering a safe alternative to testing new control strategies directly on human subjects in physical settings. However, a significant challenge remains in ensuring that these digital twins accurately represent the real human body and its interactions with the environment, which is commonly referred as a sim-to-real gap \cite{salvato2021crossing, sudhakar2023exploring}. The fidelity of these models is vital for translating simulation results into practical applications, and ongoing research aims to bridge the gap between virtual simulations and real-world performance. \par

Significant efforts have been made within the biomechanics community to develop realistic simulation environments for human movement analysis. Among these, OpenSim \cite{delp2007opensim}, an open-source platform, is renowned for its comprehensive musculoskeletal models, enabling researchers to study joint dynamics and muscle forces. Despite its strong community support, OpenSim's relatively steep learning curve, high computational demands, and limited integration with machine learning (ML) frameworks present challenges for researchers looking to leverage data-driven approaches. Similarly, AnyBody \cite{damsgaard2006analysis}, a proprietary system, offers detailed musculoskeletal modeling with robust industrial applications in ergonomics and orthopedics. However, its high cost and restricted accessibility hinder widespread adoption. Recent advancements, such as MyoSuite \cite{caggiano2022myosuite}, have begun bridging these gaps by offering open-source musculoskeletal environments built on the multi-contact physics engine MuJoCo \cite{todorov2012mujoco}. MyoSuite combines anatomical fidelity with seamless integration into widely used machine learning frameworks, making it well-suited for tasks such as musculoskeletal motor control using reinforcement learning; however, it is relatively underdeveloped compared to more established tools. These tools collectively demonstrate a spectrum of capabilities, catering to various research needs and priorities. \par

Accurately recreating human movements in virtual models is vital for analyzing gait biomechanics, as realistic simulation environments alone are insufficient. Replaying motion capture data alone is insufficient because it does not account for variations in human morphology (e.g., differences in height), tasks, and activities \cite{corazza2010markerless}. In addition, locomotive strategies often vary between individuals. To address these challenges, researchers have developed control strategies for virtual skeletal humanoids to mimic human gait biomechanics. Techniques such as imitation learning and reinforcement learning have been employed to simulate human locomotion in virtual environments \cite{haarnoja2018soft, peng2018deepmimic}. However, these methods focus on training agents to replicate kinematic profiles for a single locomotor task or a single target speed \cite{dembia2020opensim, dashkovets2024reinforcement}. Such approaches lack generalizability for different tasks, speeds, and subjects, significantly limiting their applicability. This constraint undermines the utility of virtual models, restricting their use to narrow, task-specific scenarios rather than to the diverse applications that they are intended to support. Although there are ongoing efforts to train a single policy that can generalize across different speeds within specific activities, these studies typically employ agents with limited DOF compared to musculoskeletal models, reducing their potential for real-world interpretability \cite{haarnoja2018soft, peng2018deepmimic, peng2018variational}. \par

In this work, we trained and validated a skeletal humanoid agent using imitation learning, capable of generalizing across various walking speeds using a high-fidelity physics engine. The key contributions of our study are as follows: (1) we developed a robust pipeline for generating synthetic lower-limb motion data derived from real-world lower-extremity gait biomechanics dataset, (2) using this synthetic dataset, we trained an agent with adversarial imitation learning that is capable of a natural and stable locomotion, (3) we performed an ablation study incorporating curriculum learning and an additional speed-based reward term to evaluate the agent’s generalizability across speeds, and (4) we conducted comprehensive statistical analyses to compare the synthetic motion data-based agent's resulting kinematics and the original gait biomechanics dataset. Our work represents a significant step toward developing a digital twin that replicates human lower-limb gait biomechanics, with potential applications in biomechanical research, the design of wearable assistive devices, and rehabilitation. Our implementation of the proposed pipeline is publicly available at: \url{https://github.com/MetaMobilityLabCMU/speed-adaptive-agent}.      \par

\section{Methods}
This study aims to develop a virtual locomotion agent capable of adapting to varying walking speeds while exhibiting gait profiles that closely resemble those of its biological counterpart. We designed a framework to train a speed-adaptive agent using imitation learning to replicate expert demonstrations derived from an open-source biomechanics dataset \cite{camargo2021comprehensive} (Fig. \ref{fig:overview}). This section outlines the simulation environment and skeletal model utilized, the training pipeline using our synthetic data generator, the learning algorithm via curriculum strategies, and the statistical analyses to compare the agent's performance against a baseline benchmark.

\subsection{Environment and Task}
We used LocoMuJoCo \cite{al2023locomujoco} as the simulation environment to train a walking skeletal agent. LocoMuJoCo is designed to enable researchers to train and evaluate ML-based agents for locomotion tasks in realistic and complex environments. LocoMuJoCo was chosen for three primary reasons: (1) it incorporates skeletal models with a high number of DOF, closely resembling human anatomy and biomechanics, (2) it provides native support for integrating ML software packages and algorithms, and (3) it employs the high-fidelity physics simulator MuJoCo \cite{todorov2012mujoco} in the backend, enabling realistic interaction with the environment by incorporating multi-contact dynamics.

Among the skeletal humanoid models available in LocoMuJoCo, we selected the one tailored to our study's focus on locomotion. LocoMuJoCo provides muscle and skeletal models and has four different categories of age groups. Specifically, we utilized an adult-sized skeletal model in which joint torques are directly actuated, bypassing the complexity of muscle contraction and the higher DOF present in muscle-actuated models. This model preserves the relevant joints (i.e., action and observation spaces) for walking while simplifying or omitting other components of the body (e.g., upper limb). The model, adapted from \cite{hamner2010muscle}, represents an adult human with a total body mass of 86.62 kg. The upper body was fixed, and the foot structure is represented as a rigid box, omitting the subtalar and metatarsophalangeal joints. The observation space comprises 36 dimensions, which capture the angles and angular velocities of the lumbar, hip, knee, and ankle joints, as well as the vertical position and velocity of the pelvis. The action space includes 13 dimensions, controlling the torques applied to the lumbar, hip, knee, and ankle joints. These abstractions maintain essential locomotion dynamics while improving computational efficiency. \par

The LocoMuJoCo environment provides a versatile platform for studying locomotion tasks. LocoMuJoCo offers two locomotion tasks: walking and running. For the scope of our study, we selected the walking task. Additionally, the environment provide datasets for each task, and we utilized the dataset consists of a single subject walking for 15 minutes at a speed of 1.25 m/s, post-processed from a motion capture system \cite{al2023locomujoco}. 

\subsection{Synthetic Motion Data Generator}
We developed a synthetic motion data generator to approximate biomechanics across a diverse range of walking speeds. Humans adapt their walking patterns to environmental conditions (e.g., terrain context) and intentions (e.g., walking faster) by adjusting stride length, gait, and balance. Capturing these variations is crucial to creating a digital twin capable of producing human-like locomotion in everyday scenarios. However, most existing datasets only capture discrete conditions within a task (e.g., specific speeds or slope inclines), primarily due to the limitations of generating such conditions in lab settings with human subjects \cite{hu2018benchmark, lencioni2019human,moreira2021lower}. While collecting multi-speed motion capture data is feasible, it requires significant time and resources. To address these challenges, we designed the generator to produce synthetic yet biomechanically natural data, enabling the creation of realistic gait trajectories for training a virtual agent. \par 

The synthetic motion data generator leverages an open-source lower-limb biomechanics dataset \cite{camargo2021comprehensive} that provides treadmill walking data from 22 subjects, covering speeds from 0.5 m/s to 1.85 m/s in 0.5 m/s increments. The dataset includes detailed lower-limb joint (i.e., hip, knee, and ankle) kinematics and kinetics, offering a comprehensive biomechanical representation of human gait. Using this dataset, we aggregated and averaged gait profiles across subjects within a gait cycle (heel strike to heel strike of the right leg) at each speed. We created synthetic gait kinematics by fitting a linear model to the open-source biomechanics dataset averaged across subjects. By applying the linear model to the LocoMuJoCo's single-speed walking dataset, we generated artificial data that vary across speeds. The use of synthetic kinematics instead of the open-source dataset ensured adherence to the model's physical morphology while maintaining realistic biomechanical correlations across varying speeds. Specifically, we first uniformly distributed 21 points within a gait cycle from the open-source biomechanics dataset, and then a linear model was fitted to each point as a function of walking speed. Synthetic velocity data were derived by differentiating the synthetic position data. The quality of the synthetic data was evaluated by comparing the gait profiles of the synthetic dataset to the open-source biomechanics dataset. The root mean square error (RMSE) and  the coefficient of determination ($\textit{R}^2$) were calculated for the hip, knee, and ankle joint kinematics, which were averaged over all gait cycles, providing a quantitative evaluation of the generated joint trajectories. \par 

\subsection{Training Speed-Adaptive Imitation learning Agent}
We used Variational Adversarial Imitation Learning (VAIL) to train a control policy of a skeletal humanoid agent to walk under varying speed conditions \cite{peng2018variational}. VAIL follows the general structure of a Generative Adversarial Network 
 (GAN), where the generator acts as the policy $\pi$ and the discriminator $D$ distinguishes between expert and generated trajectories \cite{goodfellow2020generative}. The optimization objective of GAN was represented by the following min-max formulation:
\begin{equation}
    \max _\pi \min _D \quad \mathbb{E}_{\mathbf{s} \sim \pi^*(\mathbf{s})}[-\log (D(\mathbf{s}))]+\mathbb{E}_{\mathbf{s} \sim \pi(\mathbf{s})}[-\log (1-D(\mathbf{s}))]
\end{equation}
The unique feature of VAIL lies in its use of an encoder to map both the generator's output and the expert demonstrations to a latent space representation before passing them to the discriminator, which acts as a regularization mechanism. This approach refines the discriminator's feedback (gradients) to the generator, guiding the generator to ignore irrelevant cues and focus on minimizing its most significant differences from the expert demonstrations. \par
To facilitate the training of a speed-adaptive imitation learning agent, we incorporated the target speed into the observation space as a proxy of the human intent of desired walking speed, allowing the agent to dynamically adapt to varying speed commands. The agent processed the skeletal model's kinematics as inputs and returned joint torques as actions. The reward function was defined as follows
\begin{equation}
    r = (1 - \lambda)(-\log(1 - \frac{1}{1+\exp(-D(\mathbf{z}))}))) + \lambda r_{speed}
\end{equation}
\begin{equation}
    \mathbf{z} \sim E(\mathbf{z}|s, a)
\end{equation}
where $D(\mathbf{z})$ is the discriminator output that measures the closeness of state and action pair of the agent to the expert demonstration, $\mathbf{z}$ is the encoding sampled from the latent space representation by the encoder, and $r_{speed}$ is the speed reward which captures the difference between the target and simulated velocity. Specifically, speed reward was defined as the exponential of the negative error squared of the target speed and the current speed of the agent's center of mass (COM) velocity in the Sagittal plane. These two rewards were adjusted during training with a weighting ratio $\lambda$. This formulation allowed the agent to match its gait profiles with the expert demonstration (i.e., synthetic motion data) while simultaneously tracking the target speed. Trust Region Policy Optimization \cite{schulman2015trust} was used to optimize the policy with reward as the objective. We trained the agent for 4000 epochs with 5000 environment steps per epoch. The policy was updated every 1000 steps and the discriminator was updated every 3000 steps. \par 

\subsubsection{Curriculum learning}
Curriculum learning specifies the order of training data, which can improve the robustness and generalization of the model. In this study, we tested two curriculum approaches to train a walking agent: a random curriculum, where target speeds were randomly selected at the start of every epoch, and a progressive curriculum, where the target speed gradually cycled back and forth across the range of speeds. The range of target speeds were capped at the speeds available in the dataset, from 0.65 m/s to 1.85 m/s.

\subsubsection{Balancing reward contributions}
We also studied the balance between two reward streams: (1) speed rewards that represent how well the agent maintains the specified target speed and (2) discriminator rewards, which reflect how closely the agent matches the biomechanics in the dataset. We tested the reward ratios of discriminator reward to speed reward, ranging from 0 to 1 in increments of 0.1 to determine the optimal balance for achieving the best performance.

\subsubsection{Baseline agent}
To evaluate our speed-adaptive agent training pipeline, from synthetic data generation to training with imitation learning, we configured a baseline agent. This baseline agent had limited access to multi-speed data from our synthetic generator, but was otherwise trained under identical conditions to the agent utilizing the full training pipeline with the optimal setting. Although the baseline agent had access to different speed information during the training process, this was only through the speed-based reward and the target speed state.

\subsection{Model Evaluation and Analysis}
To evaluate a fully trained agent against ground-truth kinematic data and its generalizability across different speeds, we conducted five simulation experiments. In each trial, the agent was evaluated over three episodes at each target speed, ranging from 0.65 to 1.85 m/s in 0.1 m/s increments. Each episode lasted for a maximum of 1000 time steps or until the agent fell, whichever occurred first. During these evaluations, we collected time series data of joint kinematics for analysis. The collected data were segmented into individual gait cycles by identifying the local minimum peaks at the hip joint in the Sagittal plane. To ensure reliable gait segmentation, we discarded gait cycle data that had a stride duration less than 60 time steps (0.6 seconds).

We evaluated the agent's performance by calculating the RMSE of the averaged differences in the hip, knee, and ankle joint angles between the synthetic dataset and the generated gait profiles of the trained agent across all speed conditions. We used $\textit{R}^2$ to assess how closely the agent's predicted joint angle profiles matched the dataset. Additionally, we measured the agent's COM velocity and compared that to the target speed we prescribed during evaluation. The optimal configuration of the curriculum strategies and reward ratios was identified by the aggregated $\textit{R}^2$ values, comparing synthetic data to the agent's output, across joint angles and speeds. We prioritized $\textit{R}^2$ because RMSE values, expressed in different units (degrees and m/s), complicated cross-experiment comparisons. \par 

To compare the agent's performance between the baseline and optimal settings, we performed statistical analyses using a paired sample t-test, with a significance level of 0.05. We also evaluated the agent's ability to track dynamic target speeds. During this test, we compared the average measured speed over 10 episodes of 50-second inference with a time-varying sinusoidal target speed (i.e., chirp signal), where the frequency linearly increased from 0.01 Hz to 0.05 Hz.

\section{Results}
\subsection{Synthetic Motion Data Generator}
\begin{figure}[!t]
    \centering
    \includegraphics[width=\linewidth]{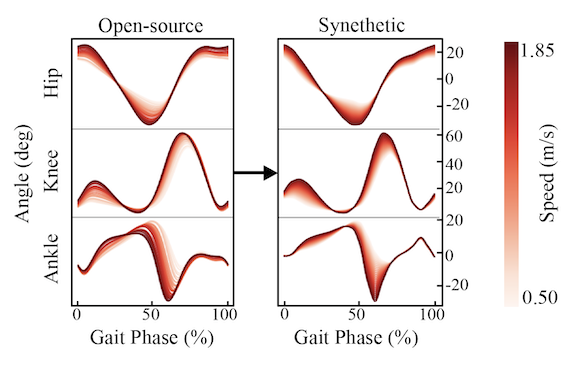}
    \caption{Comparison of joint kinematics between open-source biomechanics and synthetic data across varying speeds. A linear model was fitted to joint kinematics from an open-source gait biomechanics dataset. This model was then used to generate synthetic kinematic profiles during training.}
    \label{fig:linear_model_result}
\end{figure}

The kinematic gait profiles (i.e., joint angles) generated by the linear model demonstrated consistent variations and shapes across speeds for each lower-limb joints: hip, knee, and ankle (Fig.~\ref{fig:linear_model_result}). Across all lower-limb joints, the synthetic data resulted in an RMSE of 8.10$\pm$0.16\degree and an $\textit{R}^2$ of 0.59 compared to the biomechanical ground truth.

\begin{figure}[!b]
    \centering
    \includegraphics[width=\linewidth]{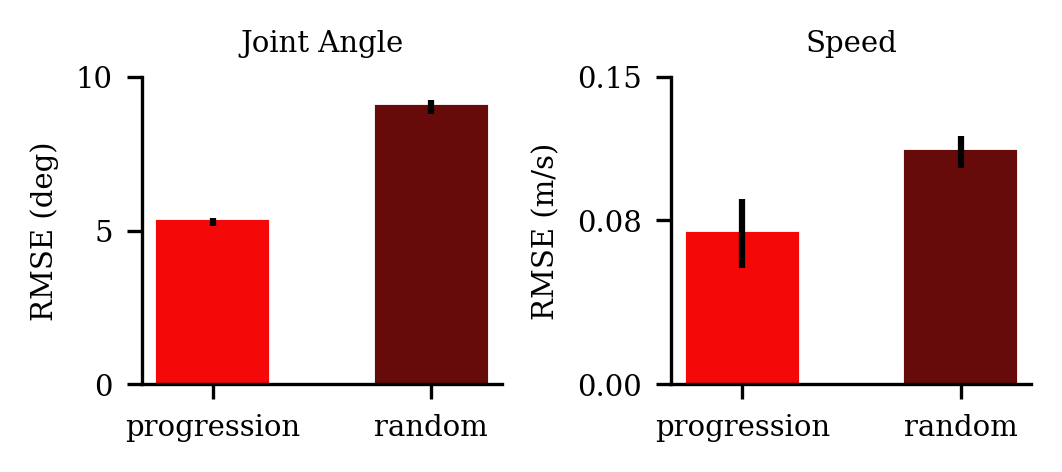}
    \caption{Effect of curriculum learning on the walking agent's performance during training. The red plot represents the case for changing the target speed in a progressive way and the brown plot stands for a random case at every training epoch. Error bars indicate $\pm$1 standard deviation.}
    \label{fig:effect_curriculum}
\end{figure}

\subsection{Effect of Curriculum Learning}
We evaluated the differences in the joint angles between the synthetic data and the trained agent at various speeds using two curriculum strategies. In addition, we analyzed the differences between the target and the simulated COM speeds achieved by the agent under these conditions. The progressive curriculum strategy resulted in a 41.48$\pm$3.06\% reduction in RMSE and a 47.77$\pm$4.37\% increase in $\textit{R}^2$ for joint angles compared to the random strategy (Fig.~\ref{fig:effect_curriculum}). Similarly, the progressive curriculum strategy achieved a 35.11$\pm$16.46\% lower RMSE and a 5.67$\pm$2.31\% higher $\textit{R}^2$ for speed tracking compared to the random strategy.

\subsection{Effect of Reward Contributions}
\begin{figure}[!b]
    \centering
    \includegraphics[width=\linewidth]{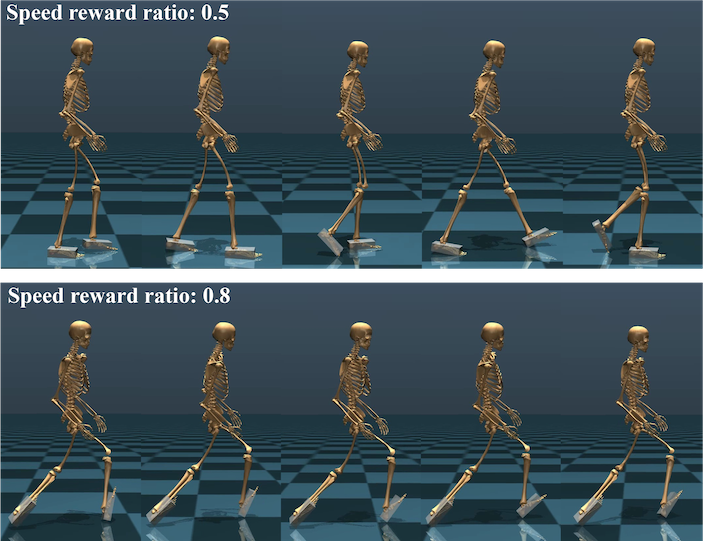}
    \caption{Representative walking agent with optimal (top) and suboptimal (bottom) settings. Suboptimal agent (e.g., higher ratio of speed reward) exhibited abnormal gait patterns, such as irregular limb coordination, exaggerated dorsiflexion, and asymmetric range of motion.}
    \label{fig:snapshots}
\end{figure}

We evaluated the joint angle differences between the synthetic dataset and the trained agent at various speeds under different ratios of two reward components: discriminator and speed (Table \ref{tab:effect_ratio_1}). In addition, we evaluated the discrepancies between the target and the simulated COM speeds of the agent under these conditions. These differences were quantified using RMSE and $\textit{R}^2$. Speed reward ratios of 0.4 and above 0.5 were excluded from our analysis due to the absence of viable samples after filtering gait cycles with values below 0.6 seconds, which exhibited unnatural gait patterns (Fig. \ref{fig:snapshots}). Among the included ratios, a speed reward ratio of 0.1 resulted in the lowest RMSE and the highest $\textit{R}^2$ for joint angle differences, while the 0.5 ratio achieved the lowest RMSE and highest $\textit{R}^2$ for speed comparisons.

\renewcommand{\arraystretch}{0.8} 
\setlength{\tabcolsep}{3pt}
\begin{table}[!h]
    \caption{Effect of weighting of the reward contributions.}
    \begin{center}
    \scriptsize\setcellgapes{2pt}\makegapedcells
    \begin{tabularx}{\linewidth}{|c|c|>{\centering\arraybackslash}X|>{\centering\arraybackslash}X|>{\centering\arraybackslash}X|>{\centering\arraybackslash}X|}
    \hline
    \multicolumn{2}{|c|}{\tiny Reward Ratio} & \multicolumn{2}{c|}{\tiny Target vs. Measured Speed} & \multicolumn{2}{c|}{\tiny Ground Truth vs. Measured Joint Angle} \\
    \hline
    \tiny Discriminator & \tiny Speed & \tiny RMSE (m/s) & $\tiny \textit{R}^2$ & \tiny RMSE (\degree) & $\tiny \textit{R}^2$\\
    \hline
    1 & 0 & 0.51$\pm$0.02 & -0.85$\pm$0.14 & 5.42$\pm$0.16 & 0.80$\pm$0.01\\
    \hline
    0.9 & 0.1 & 0.19$\pm$0.01 & 0.73$\pm$0.02 & \textbf{3.86$\pm$0.07} & \textbf{0.90$\pm$0.01}\\
    \hline
    0.8 & 0.2 & 0.13$\pm$0.01 & 0.88$\pm$0.01 & 4.97$\pm$0.06 & 0.85$\pm$0.01\\
    \hline
    0.7 & 0.3 & 0.08$\pm$0.02 & 0.95$\pm$0.02 & 5.38$\pm$0.15 & 0.81$\pm$0.01\\
    \hline
    0.6 & 0.4 & 0.07$\pm$0.01 & 0.96$\pm$0.01 & - & -\\
    \hline
    0.5 & 0.5 & \textbf{0.06$\pm$0.01} & \textbf{0.98$\pm$0.01} & 5.24$\pm$0.08 & 0.83$\pm$0.01\\
    \hline
    \end{tabularx}
    \end{center}
    \label{tab:effect_ratio_1}
    \makebox[\linewidth][c]{\footnotesize{\parbox{\linewidth}{
    Effect of varying speed and discriminator reward ratios during training. Results with bold fonts indicate the best walking performance for each evaluation criteria (i.e., target speed and kinematic profile).   

    }}}\\
\end{table}

\begin{figure}[!t]
    \centering    \includegraphics[width=\linewidth]{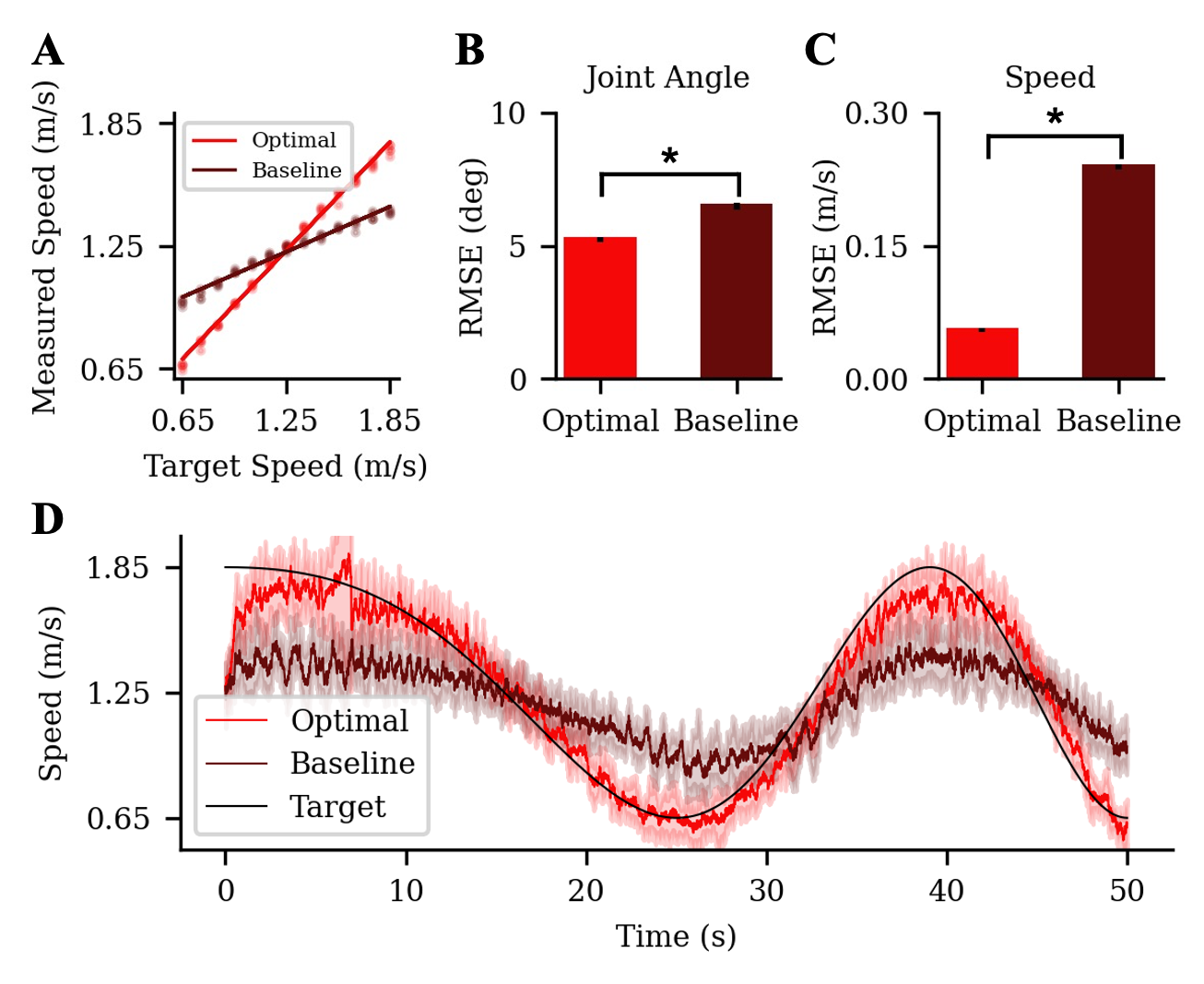}
    \caption{Walking agent performance with optimal and baseline settings. (A) Target speed tracking performance across varying conditions. Tracking error for (B) joint angles and (C) COM speed relative to ground-truth data. (D) Walking agent's adaptability to dynamically changing walking speeds. Error bars and shaded regions indicate $\pm$1 standard deviation, and asterisks indicate statistical significance ($\textit{p}<$ 0.05).
    }
    \label{fig:target_vs_actual}
\end{figure}
\begin{figure*}[!t]
    \centering
    \includegraphics[width=\linewidth]{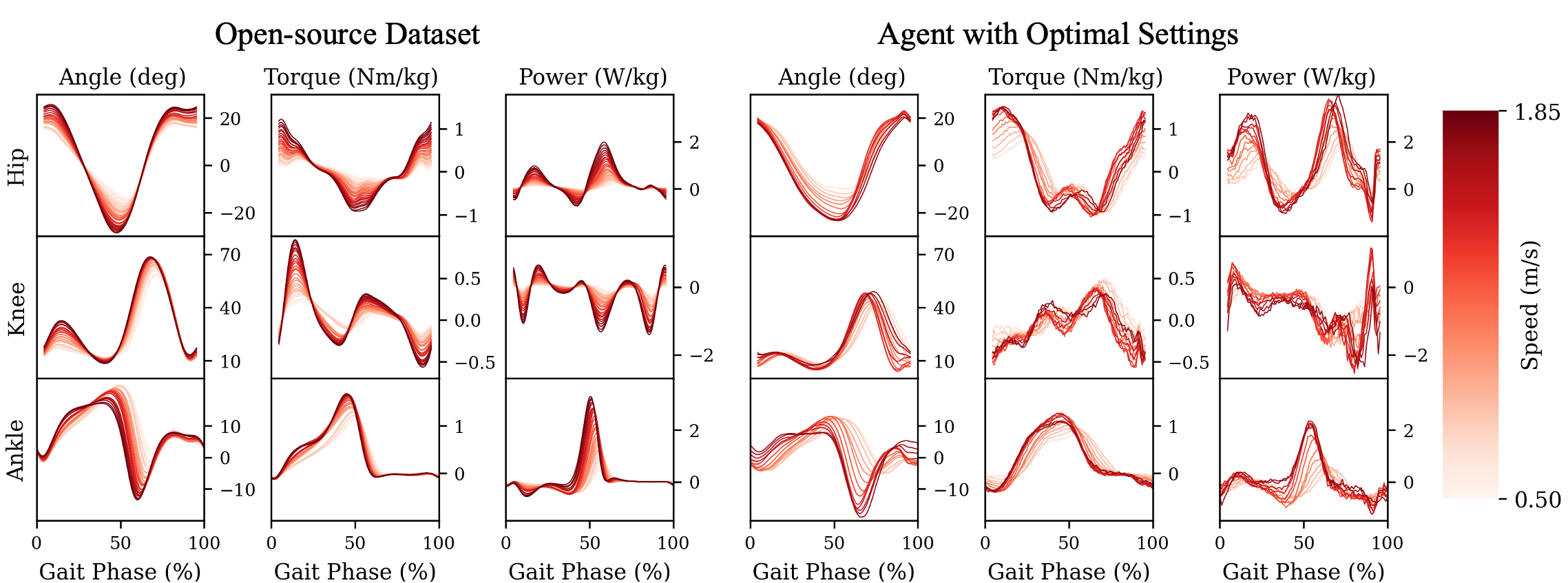}
    \caption{Gait biomechanics comparison between the open-source dataset (left) and our optimal walking agent (right) across varying speeds. Each curve from the open-source dataset represents the average gait across all subjects in the dataset, while each curve from the optimal walking agent represents the average gait across 50 episodes of evaluation. Torque and power values were normalized by body mass.}
    \label{fig:biomechanics}
\end{figure*}
\subsection{Optimal Agent Settings}
Based on the aggregated $\textit{R}^2$ values, the best-performing agent utilized a speed reward ratio of 0.5, a discriminator ratio of 0.5, and a progressive curriculum strategy. The agent with optimal settings significantly outperformed the baseline benchmark in both speed and joint angle metrics ($\textit{p}<$0.05). Specifically, the optimal agent achieved an RMSE of 0.06$\pm$0.01 m/s and 5.24$\pm$0.09°, and an $\textit{R}^2$ of 0.98$\pm$0.01 and 0.83$\pm$0.01 for speed and joint angles, respectively. In contrast, the baseline agent resulted in an RMSE of 0.24$\pm$0.01 m/s and 6.49$\pm$0.12°, and an $\textit{R}^2$ of 0.59$\pm$0.01 and 0.70$\pm$0.02 for speed and joint angles, respectively. Lastly, the optimal agent outperformed the baseline in tracking dynamically changing speeds, accurately matching targets across all frequency ranges. In contrast, the baseline struggled to follow the target speeds, only approximating the rough trend (Fig. \ref{fig:target_vs_actual}).

\section{Discussion and Conclusion}
In this work, we presented a pipeline for optimizing a speed-adaptive walking agent using synthetic motion data and adversarial imitation learning, where our optimal agent achieved a stable, biomechanically plausible locomotion. The agent with optimal settings outperformed the baseline, highlighting the effectiveness of our pipeline. This included the synthetic data generator, underscoring the importance of speed-dependent data for producing biologically plausible gait while achieving target speeds. Furthermore, an evaluation of curriculum learning strategies showed that progressive speed transitions significantly outperformed random shuffling. We attribute this to frequent and abrupt changes in the target speed under the random strategy, which likely disrupted the agent’s ability to establish a stable walking pattern. In contrast, the progressive curriculum facilitated a smoother transition between speeds, allowing enough time for the agent to develop stable walking dynamics without falling. Compared to prior methods on imitating human walking in simulation \cite{schumacher2023natural,chan2023creation,song2021deep}, our results not only achieved comparable kinematic profiles but also biomechanically plausible joint moments and powers (Fig. \ref{fig:biomechanics}). These additional metrics provide deeper biomechanical insights, extending the analysis beyond kinematic comparison, and enhancing the framework's applicability to the real-world use case. \par

Despite these contributions, our framework has several limitations. First, the synthetic motion data generator relied on a linear model to capture trends in the variation of human gait kinematics across speeds. However, gait kinematics can exhibit nonlinear behavior at certain speeds (e.g., slow walking). This assumption of generic scaling in motion data may have led to less accurate representations of actual human biomechanics. Consequently, this limitation likely contributed to additional deviations between the kinematics from the open-source dataset and our synthetic data (Fig. \ref{fig:linear_model_result}), ultimately affecting the trained agent's gait profile (Fig. \ref{fig:biomechanics}). Moreover, the linear model was fitted to the average data of 22 subjects, which filtered out individual variations in human gait dynamics. As a result, synthetic data may not adequately reflect the unique biomechanics of real individuals. \par

We investigated the use two reward components, the discriminator and the speed reward, to enable the agent to achieve a desired speed while exhibiting natural walking patterns. However, redundancy may have existed in our reward formulation, as the COM speed was already encoded in the discriminator, which evaluated kinematics against expert data at each speed. We hypothesized that an explicit speed reward would improve accuracy and our results supported this, with the RMSE decreasing as the speed reward ratio increased (Table \ref{tab:effect_ratio_1}). However, beyond a 0.5 ratio, the agent produced unrealistic gait profiles, as shown by the RMSE trend and visual inspections (Fig.~\ref{fig:snapshots}). This limitation is mainly due to the speed reward enforcing constant COM speed across gait cycles, which conflictes with the natural cyclic variations of human gait. \par

Although our optimal agent showed reduced kinematic deviations from the ground truth compared to the baseline agent, notable discrepancies were observed when compared to real human biomechanics (Fig.~\ref{fig:biomechanics}). For example, the agent's hip torque profile showed an additional extension peak around 50\% of the gait cycle. Additionally, the agent’s knee joint torque and power profiles deviated from those of biological counterparts. These discrepancies were mainly due to the gap in expert demonstrations using our synthetic data, which only provided joint kinematics but not corresponding kinetics. Lastly, a simplified box foot model, which reduces the complexity of contact dynamics, likely contributed to the additional sim-to-real gap. These limitations show that biomechanically plausible kinematics do not always translate into accurate interactions or contact physics, ultimately leading to differences in joint kinetics.

In future work, our aim is to develop a more representative human gait reward function to facilitate training schemes that produce a walking agent that is biomechanically plausible both kinematically and kinetically. Specifically, we plan to design rewards that assess the complete kinematic profile across the entire gait cycle, rather than evaluating it at individual time steps. This approach will address the previously discussed limitations of redundancy in the speed reward by dynamically assigning appropriate COM target speeds throughout the gait cycle. Additionally, this enhanced reward function will incorporate lower-limb biomechanical metrics such as stride length, cadence, and stability, further improving the agent's ability to replicate human-like gait patterns. \par
 
There are several exciting directions from our work. One is to extend the agent’s task beyond level-ground walking to more complex locomotor activities such as sit-to-stand and stair descent. Another avenue is to utilize the walking agent for tuning control policies of wearable robots, such as exoskeletons. By incorporating real-world exoskeleton hardware into the simulation environment, we can rapidly iterate and test various control strategies without the need for human subject testing. Minimizing the sim-to-real gap will enable direct transfer of control parameters to physical systems or fine-tuning policies with minimal real-world training data. This approach is particularly advantageous for robotic prostheses, where robust control strategies are critical to ensure user safety and prevent potentially harmful failures. \par

\bibliographystyle{ieeetr}
\bibliography{references}

\begin{thebibliography}{10}

\bibitem{satariano2012mobility}
W.~A. Satariano, J.~M. Guralnik, R.~J. Jackson, R.~A. Marottoli, E.~A. Phelan, and T.~R. Prohaska, ``Mobility and aging: new directions for public health action,'' {\em American journal of public health}, vol.~102, no.~8, pp.~1508--1515, 2012.

\bibitem{webber2010mobility}
S.~C. Webber, M.~M. Porter, and V.~H. Menec, ``Mobility in older adults: a comprehensive framework,'' {\em The Gerontologist}, vol.~50, no.~4, pp.~443--450, 2010.

\bibitem{cdc2024disability}
{Centers for Disease Control and Prevention}, ``Disability and health data system (dhds),'' 2024.

\bibitem{iezzoni2001mobility}
L.~I. Iezzoni, E.~P. McCarthy, R.~B. Davis, and H.~Siebens, ``Mobility difficulties are not only a problem of old age,'' {\em Journal of General Internal Medicine}, vol.~16, no.~4, pp.~235--243, 2001.

\bibitem{baker2006gait}
R.~Baker, ``Gait analysis methods in rehabilitation,'' {\em Journal of NeuroEngineering and Rehabilitation}, vol.~3, no.~1, p.~4, 2006.

\bibitem{delp2007opensim}
S.~L. Delp, F.~C. Anderson, A.~S. Arnold, P.~Loan, A.~Habib, C.~T. John, E.~Guendelman, and D.~G. Thelen, ``Opensim: open-source software to create and analyze dynamic simulations of movement,'' {\em IEEE transactions on biomedical engineering}, vol.~54, no.~11, pp.~1940--1950, 2007.

\bibitem{barricelli2019survey}
B.~R. Barricelli, E.~Casiraghi, and D.~Fogli, ``A survey on digital twin: Definitions, characteristics, applications, and design implications,'' {\em IEEE access}, vol.~7, pp.~167653--167671, 2019.

\bibitem{delp1990interactive}
S.~L. Delp, J.~P. Loan, M.~G. Hoy, F.~E. Zajac, E.~L. Topp, and J.~M. Rosen, ``An interactive graphics-based model of the lower extremity to study orthopaedic surgical procedures,'' {\em IEEE Transactions on Biomedical engineering}, vol.~37, no.~8, pp.~757--767, 1990.

\bibitem{ong2019predicting}
C.~F. Ong, T.~Geijtenbeek, J.~L. Hicks, and S.~L. Delp, ``Predicting gait adaptations due to ankle plantarflexor muscle weakness and contracture using physics-based musculoskeletal simulations,'' {\em PLoS computational biology}, vol.~15, no.~10, p.~e1006993, 2019.

\bibitem{sreenivasa2017optimal}
M.~Sreenivasa, M.~Millard, M.~L. Felis, K.~Mombaur, and S.~I. Wolf, ``Optimal control based stiffness identification of an ankle-foot orthosis using a predictive walking model,'' {\em Frontiers in computational neuroscience}, vol.~11, p.~23, 2017.

\bibitem{kumar2020learning}
V.~C. Kumar, S.~Ha, G.~Sawicki, and C.~K. Liu, ``Learning a control policy for fall prevention on an assistive walking device,'' in {\em 2020 IEEE International Conference on Robotics and Automation (ICRA)}, pp.~4833--4840, IEEE, 2020.

\bibitem{tran2022lightweight}
M.~Tran, L.~Gabert, S.~Hood, and T.~Lenzi, ``A lightweight robotic leg prosthesis replicating the biomechanics of the knee, ankle, and toe joint,'' {\em Science robotics}, vol.~7, no.~72, p.~eabo3996, 2022.

\bibitem{kang2019effect}
I.~Kang, H.~Hsu, and A.~Young, ``The effect of hip assistance levels on human energetic cost using robotic hip exoskeletons,'' {\em IEEE Robotics and Automation Letters}, vol.~4, no.~2, pp.~430--437, 2019.

\bibitem{salvato2021crossing}
E.~Salvato, G.~Fenu, E.~Medvet, and F.~A. Pellegrino, ``Crossing the reality gap: A survey on sim-to-real transferability of robot controllers in reinforcement learning,'' {\em IEEE Access}, vol.~9, pp.~153171--153187, 2021.

\bibitem{sudhakar2023exploring}
S.~Sudhakar, J.~Hanzelka, J.~Bobillot, T.~Randhavane, N.~Joshi, and V.~Vineet, ``Exploring the sim2real gap using digital twins,'' in {\em Proceedings of the IEEE/CVF International Conference on Computer Vision}, pp.~20418--20427, 2023.

\bibitem{damsgaard2006analysis}
M.~Damsgaard, J.~Rasmussen, S.~T. Christensen, E.~Surma, and M.~De~Zee, ``Analysis of musculoskeletal systems in the anybody modeling system,'' {\em Simulation Modelling Practice and Theory}, vol.~14, no.~8, pp.~1100--1111, 2006.

\bibitem{caggiano2022myosuite}
V.~Caggiano, H.~Wang, G.~Durandau, M.~Sartori, and V.~Kumar, ``Myosuite--a contact-rich simulation suite for musculoskeletal motor control,'' {\em arXiv preprint arXiv:2205.13600}, 2022.

\bibitem{todorov2012mujoco}
E.~Todorov, T.~Erez, and Y.~Tassa, ``Mujoco: A physics engine for model-based control,'' in {\em 2012 IEEE/RSJ international conference on intelligent robots and systems}, pp.~5026--5033, IEEE, 2012.

\bibitem{corazza2010markerless}
S.~Corazza, L.~M{\"u}ndermann, E.~Gambaretto, G.~Ferrigno, and T.~P. Andriacchi, ``Markerless motion capture through visual hull, articulated icp and subject specific model generation,'' {\em International journal of computer vision}, vol.~87, pp.~156--169, 2010.

\bibitem{haarnoja2018soft}
T.~Haarnoja, A.~Zhou, P.~Abbeel, and S.~Levine, ``Soft actor-critic: Off-policy maximum entropy deep reinforcement learning with a stochastic actor,'' in {\em International conference on machine learning}, pp.~1861--1870, PMLR, 2018.

\bibitem{peng2018deepmimic}
X.~B. Peng, P.~Abbeel, S.~Levine, and M.~Van~de Panne, ``Deepmimic: Example-guided deep reinforcement learning of physics-based character skills,'' {\em ACM Transactions On Graphics (TOG)}, vol.~37, no.~4, pp.~1--14, 2018.

\bibitem{dembia2020opensim}
C.~L. Dembia, N.~A. Bianco, A.~Falisse, J.~L. Hicks, and S.~L. Delp, ``Opensim moco: Musculoskeletal optimal control,'' {\em PLOS Computational Biology}, vol.~16, no.~12, p.~e1008493, 2020.

\bibitem{dashkovets2024reinforcement}
A.~Dashkovets and B.~Laschowski, ``Reinforcement learning for control of human locomotion in simulation,'' in {\em 2024 10th IEEE RAS/EMBS International Conference for Biomedical Robotics and Biomechatronics (BioRob)}, pp.~43--48, IEEE, 2024.

\bibitem{peng2018variational}
X.~B. Peng, A.~Kanazawa, S.~Toyer, P.~Abbeel, and S.~Levine, ``Variational discriminator bottleneck: Improving imitation learning, inverse rl, and gans by constraining information flow,'' {\em arXiv preprint arXiv:1810.00821}, 2018.

\bibitem{camargo2021comprehensive}
J.~Camargo, A.~Ramanathan, W.~Flanagan, and A.~Young, ``A comprehensive, open-source dataset of lower limb biomechanics in multiple conditions of stairs, ramps, and level-ground ambulation and transitions,'' {\em Journal of Biomechanics}, vol.~119, p.~110320, 2021.

\bibitem{al2023locomujoco}
F.~Al-Hafez, G.~Zhao, J.~Peters, and D.~Tateo, ``Locomujoco: A comprehensive imitation learning benchmark for locomotion,'' {\em arXiv preprint arXiv:2311.02496}, 2023.

\bibitem{hamner2010muscle}
S.~R. Hamner, A.~Seth, and S.~L. Delp, ``Muscle contributions to propulsion and support during running,'' {\em Journal of biomechanics}, vol.~43, no.~14, pp.~2709--2716, 2010.

\bibitem{hu2018benchmark}
B.~Hu, E.~Rouse, and L.~Hargrove, ``Benchmark datasets for bilateral lower-limb neuromechanical signals from wearable sensors during unassisted locomotion in able-bodied individuals,'' {\em Frontiers in Robotics and AI}, vol.~5, p.~14, 2018.

\bibitem{lencioni2019human}
T.~Lencioni, I.~Carpinella, M.~Rabuffetti, A.~Marzegan, and M.~Ferrarin, ``Human kinematic, kinetic and emg data during level walking, toe/heel-walking, stairs ascending/descending,'' {\em Figshare https://doi. org/10.6084/m9. figshare. c}, vol.~4494755, p.~v1, 2019.

\bibitem{moreira2021lower}
L.~Moreira, J.~Figueiredo, P.~Fonseca, J.~P. Vilas-Boas, and C.~P. Santos, ``Lower limb kinematic, kinetic, and emg data from young healthy humans during walking at controlled speeds,'' {\em Scientific data}, vol.~8, no.~1, p.~103, 2021.

\bibitem{goodfellow2020generative}
I.~Goodfellow, J.~Pouget-Abadie, M.~Mirza, B.~Xu, D.~Warde-Farley, S.~Ozair, A.~Courville, and Y.~Bengio, ``Generative adversarial networks,'' {\em Communications of the ACM}, vol.~63, no.~11, pp.~139--144, 2020.

\bibitem{schulman2015trust}
J.~Schulman, ``Trust region policy optimization,'' {\em arXiv preprint arXiv:1502.05477}, 2015.

\bibitem{schumacher2023natural}
P.~Schumacher, T.~Geijtenbeek, V.~Caggiano, V.~Kumar, S.~Schmitt, G.~Martius, and D.~F. Haeufle, ``Natural and robust walking using reinforcement learning without demonstrations in high-dimensional musculoskeletal models,'' {\em arXiv preprint arXiv:2309.02976}, 2023.

\bibitem{chan2023creation}
S.~S. Chan, M.~Lei, H.~Johan, and W.~T. Ang, ``Creation and evaluation of human models with varied walking ability from motion capture for assistive device development,'' in {\em 2023 International Conference on Rehabilitation Robotics (ICORR)}, pp.~1--6, IEEE, 2023.

\bibitem{song2021deep}
S.~Song, {\L}.~Kidzi{\'n}ski, X.~B. Peng, C.~Ong, J.~Hicks, S.~Levine, C.~G. Atkeson, and S.~L. Delp, ``Deep reinforcement learning for modeling human locomotion control in neuromechanical simulation,'' {\em Journal of neuroengineering and rehabilitation}, vol.~18, pp.~1--17, 2021.

\end{thebibliography}

\end{document}